\documentclass{article}

     \PassOptionsToPackage{round, sort, numbers, authoryear}{natbib}
     
     \usepackage[final]{neurips_2023}

\usepackage[utf8]{inputenc} 
\usepackage[T1]{fontenc}    
\usepackage{hyperref}       
\usepackage{url}            
\usepackage{booktabs}       
\usepackage{amsfonts}       
\usepackage{nicefrac}       
\usepackage{microtype}      
\usepackage{xcolor}         

\usepackage{array}
\newcolumntype{P}[1]{>{\centering\arraybackslash}m{#1}}
\usepackage{pifont}
\definecolor{deepgreen}{rgb}{0,0.6,0}
\definecolor{deepred}{rgb}{0.8,0,0}
\definecolor{deepyellow}{RGB}{246,190,0}

\usepackage[inline]{enumitem}
\setlist[itemize]{leftmargin=0.6cm,topsep=0pt,itemsep=-1ex,partopsep=1ex,parsep=1ex}

\definecolor{deepblue}{rgb}{0,0,0.6}
\hypersetup{
    final=true,
    plainpages=false,
    pdfstartview=FitV,
    pdftoolbar=true,
    pdfmenubar=true,
    bookmarksopen=true,
    bookmarksnumbered=true,
    breaklinks=true,
    linktocpage,
    colorlinks=true,
    linkcolor=deepblue,
    urlcolor=deepblue,
    citecolor=deepblue,
    anchorcolor=black}

\usepackage[pdftex]{graphicx}
\usepackage{amsmath}
\usepackage{cancel}
\usepackage{bigints}
\usepackage{stmaryrd}

\title{Diffusion-Augmented Neural Processes}

\author{%
  Lorenzo Bonito\\
  University of Cambridge\\
  \texttt{lb953@cam.ac.uk}\\
  \And
  James Requeima\\
  Vector Institute\\
  \texttt{james.requeima@gmail.com}\\
  \AND
  \hspace{-10pt}Aliaksandra Shysheya\\
  \hspace{-10pt}University of Cambridge\\
  \hspace{-10pt}\texttt{as2975@cam.ac.uk}\\
  \And
  \hspace{5pt}Richard E. Turner\\
  \hspace{5pt}University of Cambridge\\
  \hspace{5pt}\texttt{ret26@cam.ac.uk}\\
}

\begin{document}

\maketitle

\begin{abstract}
  Over the last few years, Neural Processes have become a useful modelling tool in many application areas, such as healthcare and climate sciences, in which data are scarce and prediction uncertainty estimates are indispensable. However, the current state of the art in the field (AR CNPs; \citealp{ARCNP}) presents a few issues that prevent its widespread deployment. This work proposes an alternative, diffusion-based approach to NPs which, through conditioning on noised datasets, addresses many of these limitations, whilst also exceeding SOTA performance.
\end{abstract}

\section{Introduction and Motivation}

Neural Processes are a class of meta-learning models which, by combining deep neural networks with features inspired by Gaussian processes, are able to operate in the small data regime, and also deliver reliable uncertainty estimates alongside their predictions. Since their original conception (CNPs; \citealp{CNP}), these models have garnered an increasing amount of popularity, with multiple improvements on the original formula having been developed throughout the past 5 years.

Autoregressive CNPs \citep{ARCNP} are the latest addition to (and current state of the art in) the NP Family. Albeit extremely performant, these models present three main issues: \begin{enumerate*}[label=(\roman*)] \item Whilst predictives far down the autoregressive chain do achieve high complexity, variables generated early on in the AR process are still limited to elementary Gaussian distributions \item Predictions are not consistent (i.e.~are not marginalisation- and ordering-invariant with respect to the target set $\mathcal{D}_t$) \item Deployment computational costs are prohibitive, scaling poorly in most practical applications \end{enumerate*}.

The work described in this paper proposes (Section \ref{sec:DANPs}) and, subsequently, implements and benchmarks (Section \ref{sec:experiments}) a novel, diffusion-inspired \citep{diff_2} data augmentation approach to NPs which aims to directly overcome all three of these intrinsic weaknesses of AR CNPs. The resulting Diffusion-Augmented Neural Processes are able, through clever conditioning on multiple noised data sources, to produce \textit{correlated}, highly \textit{non-Gaussian} and \textit{consistent} predictions at \textit{every} input location, whilst also attaining \textit{state-of-the-art performance} in our experiments and requiring \textit{less compute} during deployment (depending on the specific choice of architecture and on the learning task at hand).

\section{Diffusion-Augmented Neural Processes}\label{sec:DANPs}
The key idea at the basis of DANPs revolves around two main steps:
\begin{itemize}
    \item An augmented dataset is produced by generating a series of modified target fidelity levels through an AR process in which the zeroth fidelity level consists of the original target outputs, and all subsequent ones are obtained by adding Gaussian noise onto the previous level (Subsection \ref{subsec:data_aug}).
    \item  The augmented dataset is modelled by a NP model, in an AR denoising process wherein the noisiest fidelity level is first generated from the original context set $\mathcal{D}_c$, and each subsequent fidelity is then obtained by conditioning on both $\mathcal{D}_c$ \textit{and} the previously-obtained fidelities. This operation is repeated until the original, noise-free target points are recovered (Subsection \ref{subsec:model_arch_train}).
\end{itemize}

By applying this autoregressive denoising procedure, the augmented dataset is used to induce a non-Gaussian and complex distribution on the observed data. Crucially, this technique cleverly solves the first problem highlighted above for standard AR CNPs, as the first variables generated in the AR chain are, while definitely normally distributed, not actually of interest themselves. Additionally, since each fidelity level is modelled jointly, the ordering of the target inputs has no influence on the final predictions, meaning consistency is also retained, provided that the marginalisation invariance property is satisfied (which is always the case, as long as the locations of the points in the added fidelity levels are independent of the original target set).

As explained in more mathematical detail in the following, the introduced data augmentation serves the purpose of a \textit{latent variable}, which, when integrated out at test time, induces complex dependencies without the need for variational inference (unlike in the case of ``Latent'' NPs; \citealp{NP}).

One should also note that, unlike the similarly-named Neural Diffusion Processes \citep{dutordoir2023neural} (which are generative diffusion models whose denoising process is parameterised by a NP), the proposed DANPs are NP models designed to produce conditional predictives directly (without in-filling), and thus take a distinct modelling approach and employ a widely different architecture.

\subsection{Data Augmentation Through Noising}\label{subsec:data_aug}

In the regular NP setting, a task $\psi = \left(\mathcal{D}_c, \mathcal{D}_t\right)$ consists of a context set $\mathcal{D}_c = \{(x_n^{(c)}, y_n^{(c)})\}_{n=1}^{N_c}$ and a target set $\mathcal{D}_t = \{(x_n^{(t)}, y_n^{(t)})\}_{n=1}^{N_t}$, and the model learns to predict $p(\{y_n^{(t)}\}_{n=1}^{N_t}\vert \, \mathcal{D}_c, \{x_n^{(t)}\}_{n=1}^{N_t})$ through exposure to a series of such context set and target points pairings. For a DANP, an \textit{augmented} task $\widehat{\psi} = \left(\mathcal{D}_c, \mathcal{D}_t, \mathcal{D}_a\right)$ is produced by generating an \textit{auxiliary context dataset} $\mathcal{D}_a$:
\vspace{1pt}
\begin{equation}
    \mathcal{D}_a = \left(\{(x_{n, 1}^{(t)}, y_{n,1}^{(t)})\}_{n=1}^{N_{t,1}}, \, \dots,\, \{(x_{n, F}^{(t)}, y_{n,F}^{(t)})\}_{n=1}^{N_{t,F}}\right)
\end{equation}
through an autoregressive noising procedure, in which, starting from the zeroth fidelity level (the original target set), each element of $\{y_{n, f+1}^{(t)}\}_{n=1}^{N_t}$ is generated, given the corresponding element of $\{y_{n, f}^{(t)}\}_{n=1}^{N_t}$, through the expression $y_{n, f+1}^{(t)}=\sqrt{1-\beta}\, y_{n, f}^{(t)}+\epsilon_{n} \beta$ (where $\epsilon_{n} \sim \mathcal{N}(0,1)$ and $\beta$ is a noise variance hyperparameter). This procedure is exemplified (for the case of a sawtooth wave and $F=2$) in Figure \ref{fig:noised_sawtooth_dataset} below, where $\mathcal{D}_c$ is depicted in blue, $\mathcal{D}_t$ in orange, and the first and second elements of $\mathcal{D}_a$ ($\mathcal{D}_a^1$ and $\mathcal{D}_a^2$ in the following) in green and red respectively.

\begin{figure}[h]
    \centering
    \includegraphics[width = .98\textwidth]{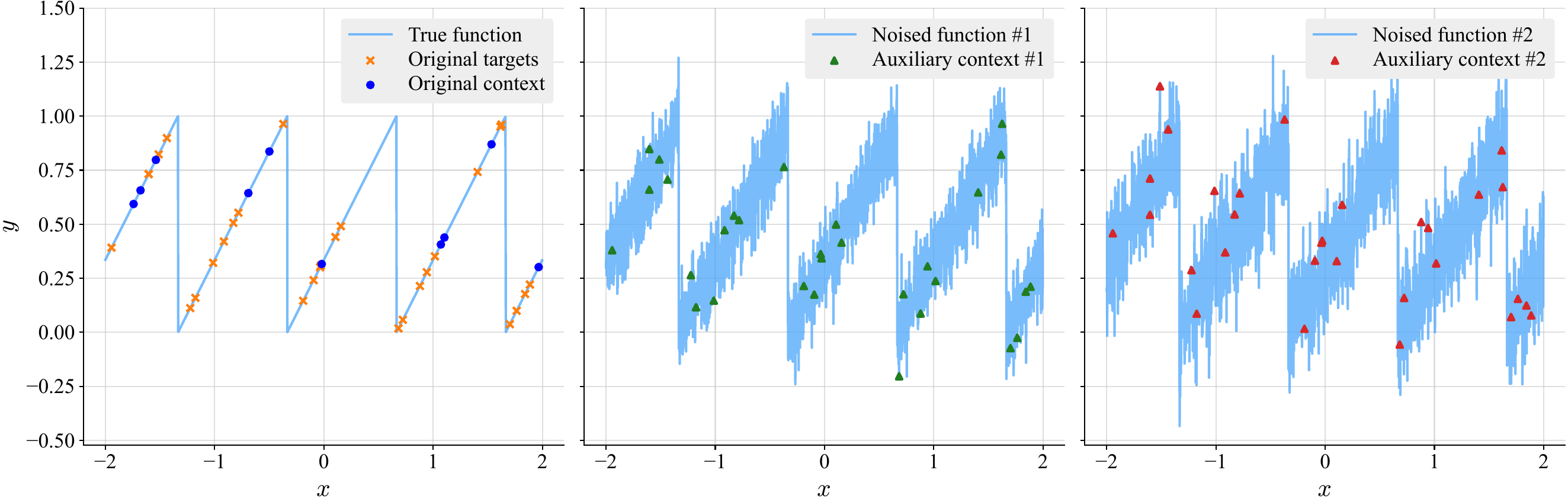}
	\caption[Noised Sawtooth Task]{A typical training task based on a sawtooth wave.}
	\label{fig:noised_sawtooth_dataset}
\end{figure}

\subsection{Model Architecture and Training}\label{subsec:model_arch_train}

\begin{figure}[h]
    \centering
    \includegraphics[width = .9\textwidth]{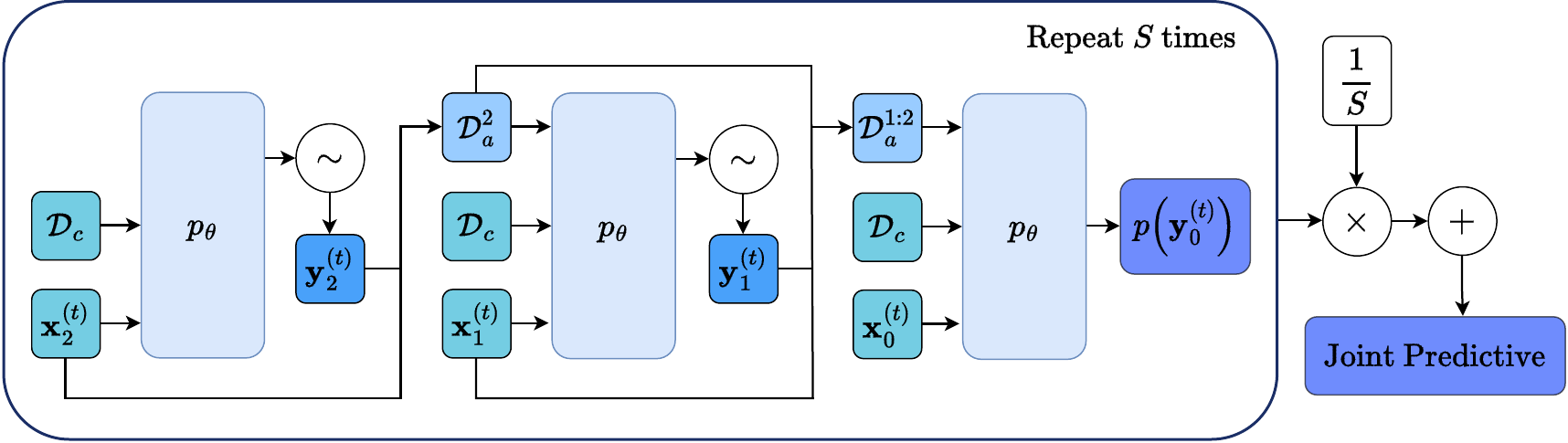}
	\caption[Diagram]{DANP architecture diagram for $F=2$.}
	\label{fig:diagram}
\end{figure}

Armed with appropriately augmented training datasets, let us now take a closer look at the modelling approach taken by DANPs. In an effort to simplify notation, we shall henceforth denote $\{x_{n, f}^{(t)}\}_{n=1}^{N_t}$ as $\mathbf{x}_f^{(t)}$ and, similarly, $\{y_{n, f}^{(t)}\}_{n=1}^{N_t}$ as $\mathbf{y}_f^{(t)}$ (where, for $f=0$, these expressions are understood to refer to the elements contained in the original target set $\mathcal{D}_t$). The joint predictive is then defined as:
\begin{equation}\label{eq:ar_danp_joint_predictive}
	\begin{aligned}
    p_\theta\!\left(\mathbf{y}_0^{(t)}\vert \,\mathbf{x}_0^{(t)},\mathcal{D}_c\right) &= \int\!p_\theta\!\left(\mathbf{y}_0^{(t)}\vert \,\mathbf{x}_0^{(t)}, \mathcal{D}_c, \mathcal{D}_{a} \right) p_\theta\!\left(\mathcal{D}_{a}\vert\,\mathcal{D}_c\right)\mathrm{d}\mathcal{D}_{a}\\[-5pt] &\approx \sum\nolimits_{s=1}^{S} \frac{1}{S}\,p_\theta\!\left(\mathbf{y}_0^{(t)}\vert \,\mathbf{x}_0^{(t)}, \mathcal{D}_c, \mathcal{D}_{a,s} \right)
    \end{aligned}
\end{equation}
where the $\approx$ sign denotes a Monte-Carlo approximation obtained by conditioning on \textit{different} instances $\mathcal{D}_{a,s}$ of the auxiliary context dataset $\mathcal{D}_a = \left(\mathcal{D}_a^{1}, \dots,\, \mathcal{D}_a^{F}\right)$. Each $\mathcal{D}_{a,s}$ is obtained by repeating (with different random states) the following autoregressive sampling procedure (all auxiliary inputs $\mathbf{x}_F^{(t)}$ through $\mathbf{x}_1^{(t)}$ are sampled independently and uniformly from the appropriate range):
\begin{equation}\label{eq:ar_procedure_D_a}
    \begin{aligned}
        \mathcal{D}_a^{\,F} = \left(\mathbf{x}_F^{(t)}, \mathbf{y}_F^{(t)}\right);& \qquad \mathbf{y}_F^{(t)} \sim \,p_\theta\!\left(\mathbf{y}_F^{(t)}\vert \,\mathbf{x}_F^{(t)}, \mathcal{D}_c\right)\\[-7pt]
        &\quad\vdots\\[-7pt]
        \mathcal{D}_a^{f} = \left(\mathbf{x}_{f}^{(t)}, \mathbf{y}_{f}^{(t)}\right);& \qquad \mathbf{y}_{f}^{(t)} \sim \,p_\theta\!\left(\mathbf{y}_{f}^{(t)}\vert \,\mathbf{x}_{f}^{(t)}, \mathcal{D}_c, \mathcal{D}_a^{f+1:F}\right)\\
    \end{aligned}
\end{equation}
where the notation $\mathcal{D}_a^{f:F}$ is employed as a shorthand for the collection $((\mathbf{x}_{f}^{(t)}, \mathbf{y}_{f}^{(t)}), \dots,  (\mathbf{x}_F^{(t)}, \mathbf{y}_F^{(t)}))$, and the tuples $(\mathbf{x}_f^{(t)}, \mathbf{y}_f^{(t)})$ are utilised in lieu of the more cumbersome $ \{(x_{n, f}^{(t)}, y_{n,f}^{(t)})\}_{n=1}^{N_t}$. To summarise the approach, the procedures set out in Equations \ref{eq:ar_danp_joint_predictive} and \ref{eq:ar_procedure_D_a} are exemplified through the diagram presented in Figure \ref{fig:diagram} below (for the case of $F=2$).

$p_\theta$ is parameterised using a NP, which, tasked with jointly modelling \textit{all} fidelity levels, is constructed to handle a context set with $F+1$ channels (1 for $\mathcal{D}_c$ and $F$ for $D_a$), and to produce $F+1$-channel target predictions given $F+1$-channel target locations. Notably, when dealing with a specific fidelity level $f'$, all target inputs $\{\mathbf{x}_f^{(t)}\}_{f\neq f'}$ are replaced by an empty tensor, and, consequently, the resulting target predictions $\{\mathbf{y}_f^{(t)}\}_{f\neq f'}$ are also empty (which is crucial to ensure that gradient computations are not affected by data that should not be available to the model at a given training instance).

\paragraph{Training}
Much like the regular NPs they aim to extend, DANPs are also trained through a maximum likelihood objective. However, since the latter present multiple layers (each dealing with a differently-noised version of the same data), a new technique needs to be introduced: \textit{task masking}. When training a given layer $f$, the augmented task $\widehat{\psi}$ should be masked to take the following form: $\widehat{\psi}_f = (\mathcal{D}_c, \mathcal{D}_t^f, \mathcal{D}_a^{f+1:F})$, where $\mathcal{D}_t^f = \mathcal{D}_a^f$ if $f\neq0$, and simply equals $\mathcal{D}_t$ otherwise (i.e.~in the case of the last/output layer). In other words, at layer $f$, the model should learn to condition on $\mathcal{D}_c$ and on all the noisier fidelity levels $\mathcal{D}_a^{f+1:F}$ to predict (as targets) the input-output pairs contained within fidelity level $f$ (so that the AR procedures detailed in Equation \ref{eq:ar_procedure_D_a} may be carried out as required during deployment). In practice, this means that, for each training batch, a layer index $f$ is first randomly selected from $\{0, \dots, F\}$, and each augmented task within the batch is then masked accordingly. Subsequently, a batched forward pass is carried out, the relevant log-likelihood $p_{\theta}(\mathbf{y}_{f}^{(t)}\vert \,\mathbf{x}_{f}^{(t)}, \mathcal{D}_c, \mathcal{D}_a^{f+1:F})$ is computed, and its gradients employed for backpropagation.

\section{Experimental Results}\label{sec:experiments}

The performance of the proposed DANP model was benchmarked through 1D regression tasks on synthetic datasets. Three functional forms were considered: sawtooth waves, square waves and samples from Gaussian processes. As evidenced by the average likelihoods presented in Table \ref{tab:results} below, DANPs easily outperform AR ConvCNPs on all three meta-datasets, and achieve the best results of all NP models on the first two (being outperformed by ConvGNPs on the GP dataset, which is unsurprising, considering GNPs are designed to directly parameterise GPs). For the sake of brevity, a more comprehensive visual breakdown of these findings is included below only for the sawtooth dataset, with further results and experiment setups being detailed in Appendices \ref{app:further_results} and \ref{app:exp_details} respectively.

\begin{table}[h]
  \caption{Average likelihoods attained by DANPs and other NP models on 1D regression tasks.}
  \label{tab:results}
  \centering
  \begin{tabular}{@{}P{2cm} P{2cm}P{2cm}P{2cm}P{2cm}@{}}
    \toprule
      & DANP & ConvCNP & ConvGNP & AR ConvCNP\\
    \midrule
    Sawtooth & \bf 82.26 {\small$\pm 2.51$} &  32.23 {\small$\pm 2.80$} & 38.02 {\small$\pm 3.36$} & 63.23 {\small$\pm 2.81$}\\
    \midrule
    Square Wave & \bf 97.16 {\small$\pm 4.90$} &  13.55 {\small$\pm 1.51$} & 10.30 {\small$\pm 1.31$} & 23.25 {\small$\pm 2.19$}\\
    \midrule
    GP & 39.52 {\small$\pm 2.88$} & 7.19 {\small$\pm 1.12$} & \bf 45.38 {\small$\pm 3.26$} & 29.11 {\small$\pm 2.29$}\\
    \bottomrule
  \end{tabular}
\end{table}

\begin{figure}[h]
    \centering
    \includegraphics[width = .98\textwidth]{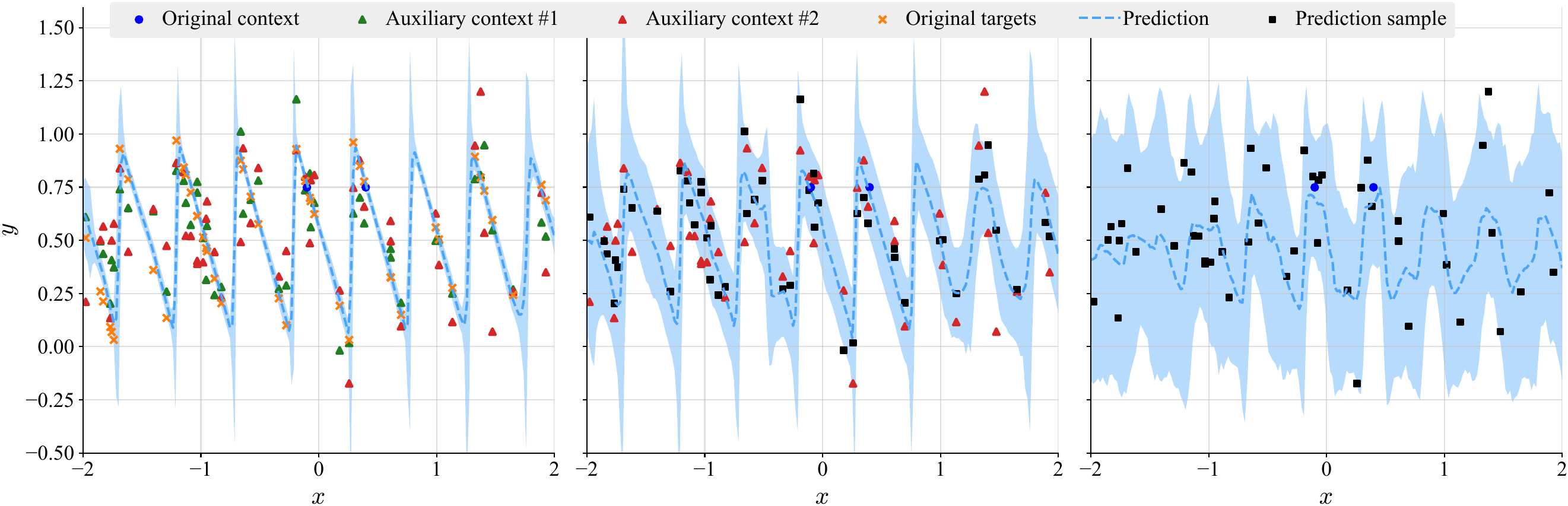}
	\caption[Noised Sawtooth Pred]{Example of DANP predictives in layers 0 (left), 1 (middle) and 2 (right) for a sawtooth benchmark task with $\lvert\mathcal{D}_c\rvert=2$. The deployment process starts in layer 2 and ends in layer 0.}
	\label{fig:noised_sawtooth_pred}
\end{figure}

\begin{figure}[h]
    \centering
    \includegraphics[width = .98\textwidth]{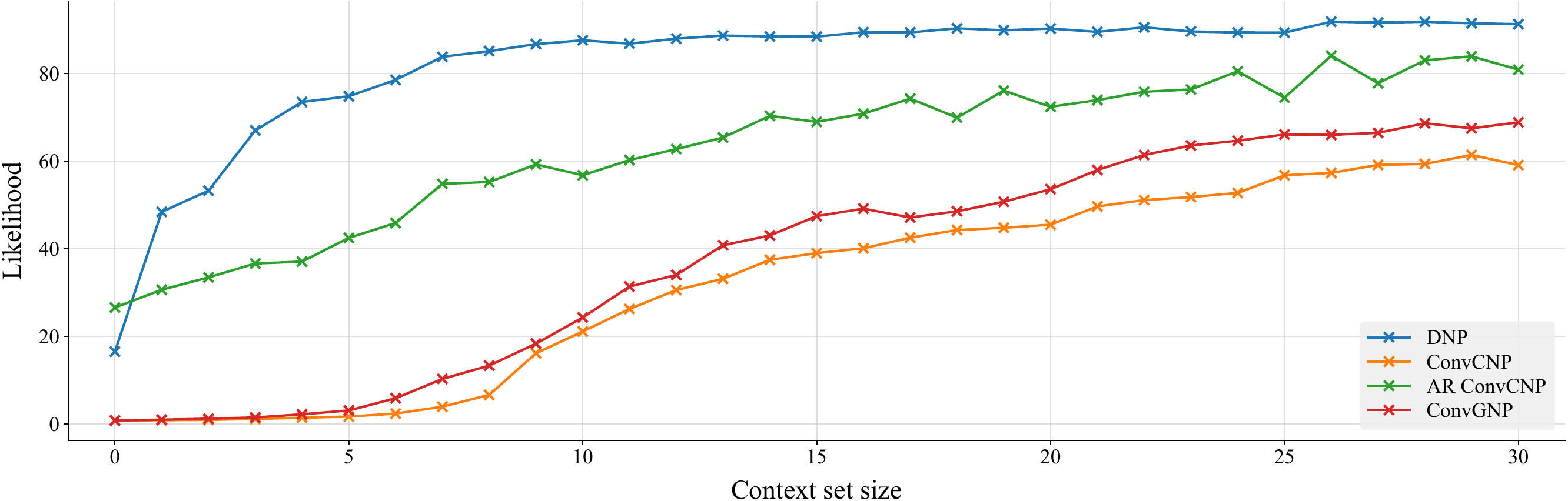}
	\caption[Noised Sawtooth Results]{Comparison of sawtooth regression DANP performance against relevant baselines. Each data point reports the average likelihood obtained across 10 tasks with a specific context size $\lvert\mathcal{D}_c\rvert$.}
	\label{fig:noised_sawtooth_results}
\end{figure}

As shown in Figure \ref{fig:noised_sawtooth_pred}, by making use of the auxiliary context sets, the DANP is able to produce especially accurate predictions (which is, in this case, particularly impressive, since only two context points are given). Additionally, the model attains higher likelihoods than its counterparts not only on tasks with average context set sizes, but across virtually all explored values of $\lvert\mathcal{D}_c\rvert$ (Figure \ref{fig:noised_sawtooth_results}).

\paragraph{Computational Costs} To tackle a task with target set size $N_t = \lvert\mathcal{D}_t\rvert$, an AR CNP performs $N_t$ forward passes (FPs), whilst a DANP with $F+1$ layers taking $S$ samples performs $S\!\left(F+1\right)$. Hence, as $N_t$ increases, the number of FPs \textit{scales linearly} in the former approach, but \textit{remains constant} in the latter. Thus, assuming a FP requires approximately the same amount of operations in both scenarios, it is computationally cheaper to deploy a DANP than an AR CNPs when predictions must be made at a large number of target locations (which happens to be the case in most practical scenarios).

\section{Conclusion}

In summary, the proposed Diffusion-Augmented Neural Processes achieve all three goals set out in the introduction, managing to not only successfully address the technical shortcomings of AR CNPs, but to also outperform them in all three 1D regression tasks examined thus far. To further confirm the validity of these findings, more regression experiments shall be conducted in the future, involving real 1D and 2D data from a variety of subject areas (climate sciences applications are especially enticing).

\section{Acknowledgements}

Aliaksandra Shysheya is supported by an EPSRC Prosperity Partnership EP/T005386/1 between the EPSRC, Microsoft Research and the University of Cambridge. Richard E. Turner is supported by Google, Amazon, ARM, Improbable and EPSRC grant EP/T005386/1.

\bibliographystyle{plainnat}
\bibliography{references}

\clearpage
\appendix
\section{Additional Results}\label{app:further_results}

\subsection{Influence of Number of Samples}

\begin{figure}[h]
    \centering
    \includegraphics[width = .98\textwidth]{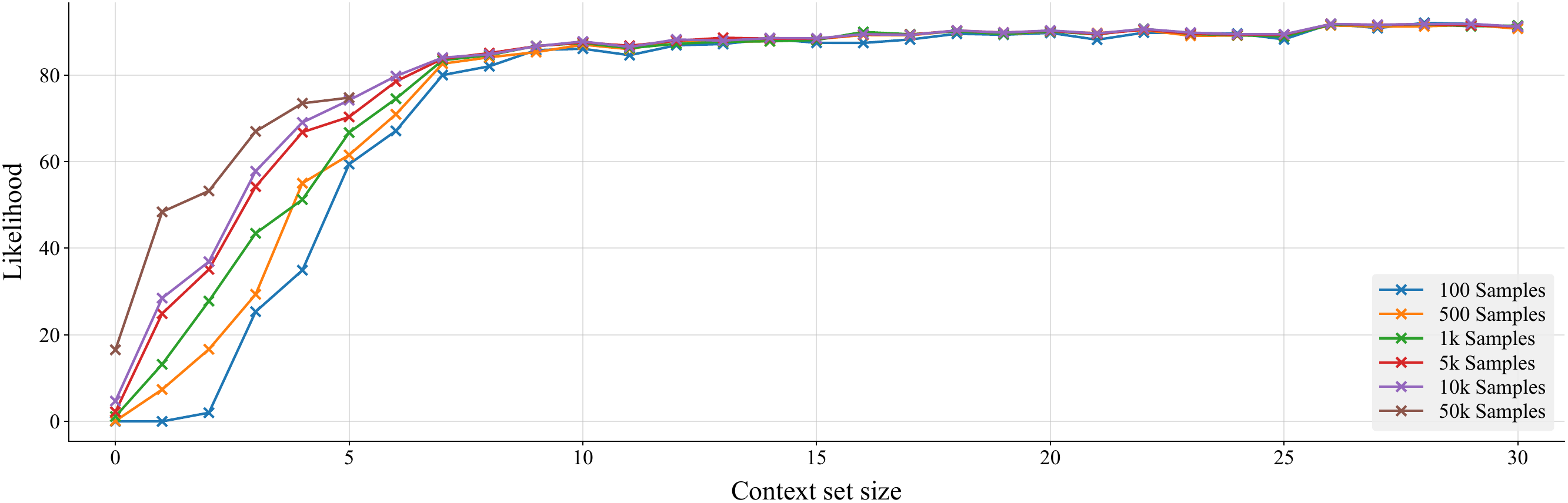}
	\caption[Number Samples Comparison]{Comparison of sawtooth regression DANP performance as the number $S$ of Monte-Carlo samples increases. Each data point reports the average likelihood obtained across 10 tasks with a specific context size $\lvert\mathcal{D}_c\rvert$.}
	\label{fig:num_samples_comp}
\end{figure}

As demonstrated in Figure \ref{fig:num_samples_comp}, increasing the number of Monte-Carlo samples (the $S$ mixture components in Equation \ref{eq:ar_danp_joint_predictive}) vastly boosts DANP performance in the low-context regime, at the price of increased computational costs.

Since, as the context set size $\lvert\mathcal{D}_c\rvert$ increases, this effect becomes less and less relevant, in an effort to attain the best possible results whilst also saving on precious compute, the data presented in Table \ref{tab:results} and Figure \ref{fig:noised_sawtooth_results} above (as well as in Figures \ref{fig:noised_square_wave_results} and \ref{fig:noised_gp_results} below) have been collected by deploying the \textit{same} models with varying values of $S$ based on context size. Namely:
\begin{itemize}
	\item For the sawtooth meta-dataset, 50k samples were drawn for $0\leq\lvert\mathcal{D}_c\rvert\leq5$, and 5k were taken for $6\leq\lvert\mathcal{D}_c\rvert\leq30$.
	\item For the square wave meta-dataset, 50k samples were drawn for $0\leq\lvert\mathcal{D}_c\rvert\leq5$, and 5k were taken for $6\leq\lvert\mathcal{D}_c\rvert\leq30$.
	\item For the GP meta-dataset, 50k samples were drawn for $0\leq\lvert\mathcal{D}_c\rvert\leq9$, and 5k were taken for  $10\leq\lvert\mathcal{D}_c\rvert\leq30$.
\end{itemize}

\subsection{Square Wave Meta-Dataset}

Similar plots to those included in the main text for the sawtooth meta-dataset are presented, for the square wave meta-dataset, in Figures \ref{fig:noised_square_wave_pred} and \ref{fig:noised_square_wave_results} below. Identical considerations to those made above still hold true.

\begin{figure}[h]
    \centering
    \includegraphics[width = .98\textwidth]{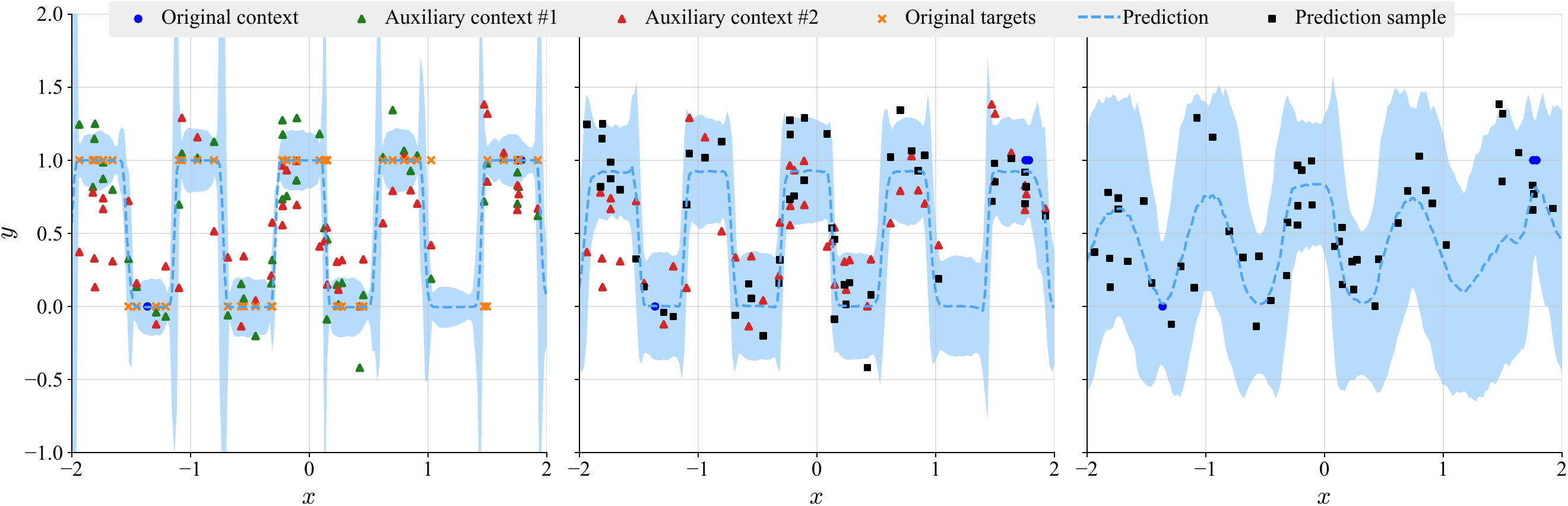}
	\caption[Noised Square Wave Pred]{Example of DANP predictives in layers 0 (left), 1 (middle) and 2 (right) for a square wave benchmark task with $\lvert\mathcal{D}_c\rvert=3$. The deployment process starts in layer 2 and ends in layer 0.}
	\label{fig:noised_square_wave_pred}
\end{figure}
\clearpage

\begin{figure}[h]
    \centering
    \includegraphics[width = .98\textwidth]{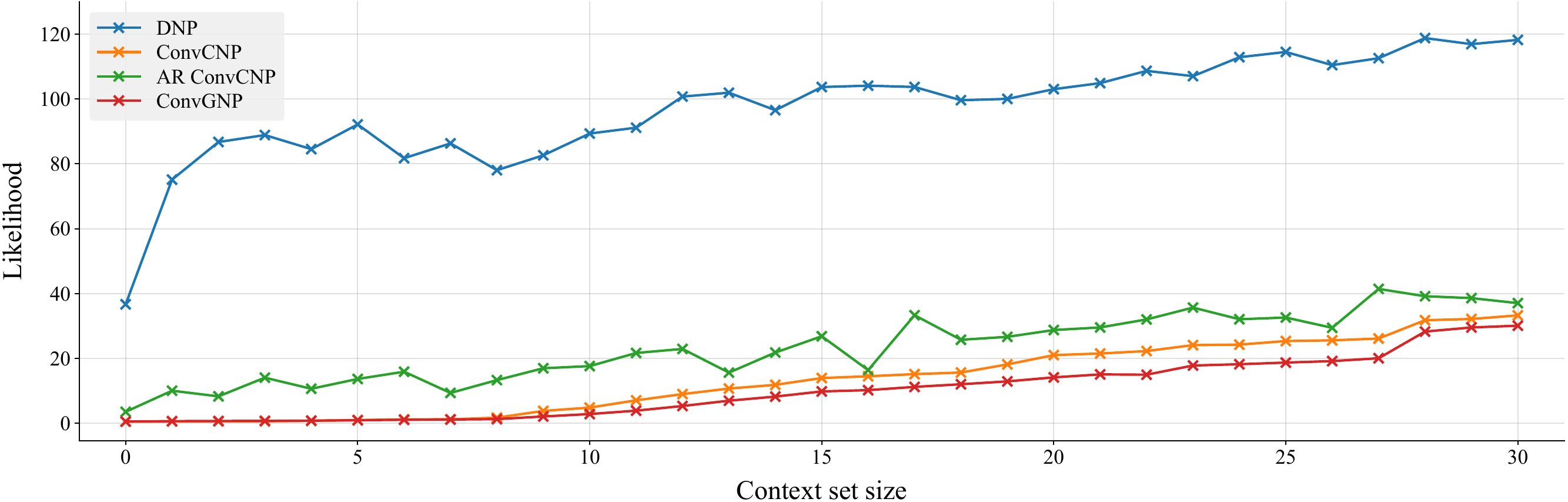}
	\caption[Noised Square Wave Results]{Comparison of square wave regression DANP performance against relevant baselines. Each data point reports the average likelihood obtained across 10 tasks with a specific context size $\lvert\mathcal{D}_c\rvert$.}
	\label{fig:noised_square_wave_results}
\end{figure}

\subsection{GP Meta-Dataset}

Similar plots to those included in the main text for the sawtooth meta-dataset are presented, for the GP meta-dataset, in Figures \ref{fig:noised_gp_pred} and \ref{fig:noised_gp_results} below. Identical considerations to those made above still hold true.

\begin{figure}[h]
    \centering
    \includegraphics[width = .98\textwidth]{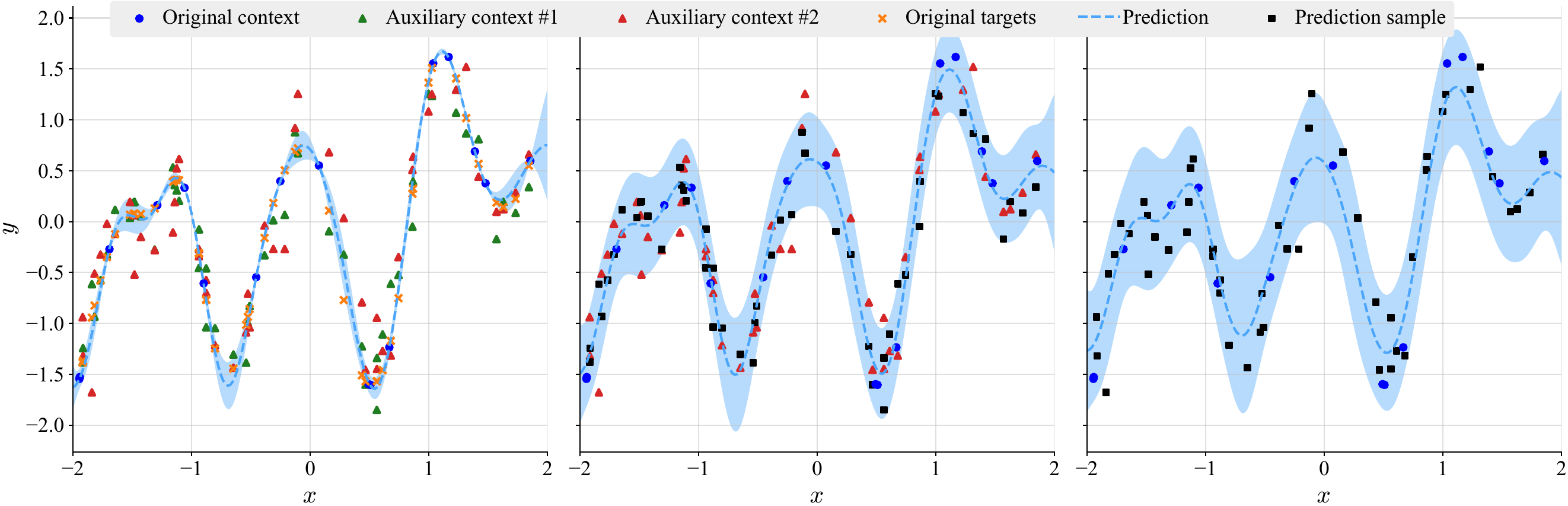}
	\caption[Noised GP Pred]{Example of DANP predictives in layers 0 (left), 1 (middle) and 2 (right) for a GP benchmark task with $\lvert\mathcal{D}_c\rvert=17$. The deployment process starts in layer 2 and ends in layer 0.}
	\label{fig:noised_gp_pred}
\end{figure}

\begin{figure}[h]
    \centering
    \includegraphics[width = .98\textwidth]{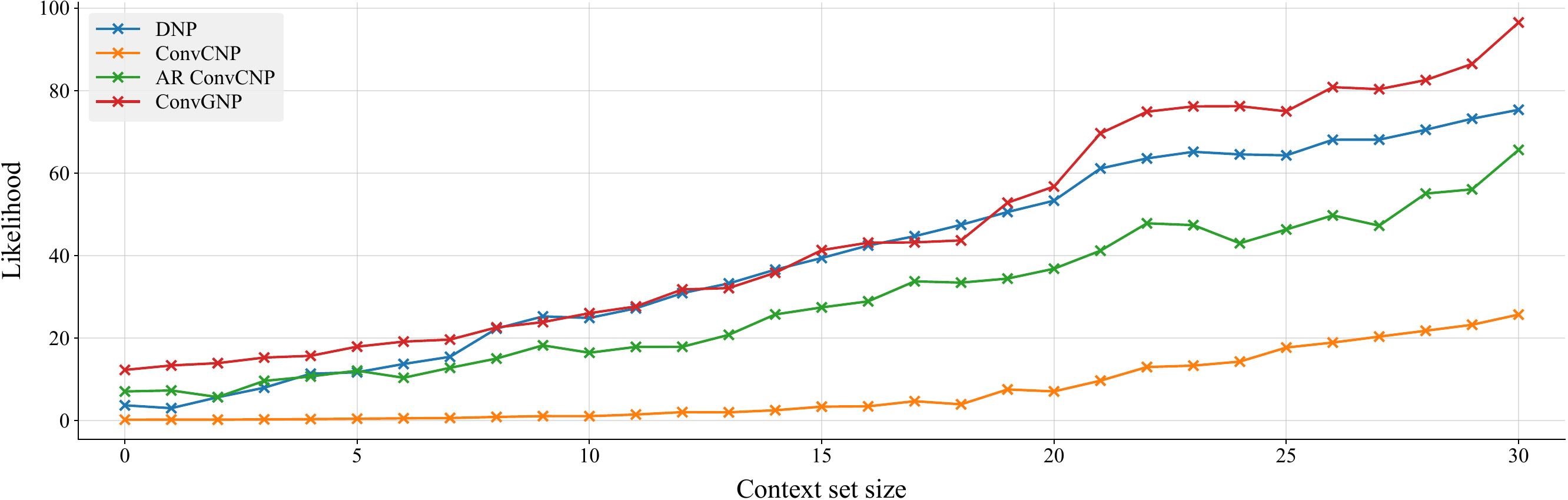}
	\caption[Noised GP Results]{Comparison of GP regression DANP performance against relevant baselines. Each data point reports the average likelihood obtained across 10 tasks with a specific context size $\lvert\mathcal{D}_c\rvert$.}
	\label{fig:noised_gp_results}
\end{figure}

\clearpage
\section{Experiments Setup}\label{app:exp_details}

\subsection{Data Generators}

To create the three meta-datasets, different functions of each form are generated as follows:
\begin{itemize}
    \item For the sawtooth dataset, each task's true function is generated by first sampling a wave frequency $\omega \sim \mathcal{U}\!\left(2,4\right)$, a direction $d$ ($-1$ or $1$ with equal probability) and a phase shift $\phi \sim \mathcal{U}\!\left(\frac{1}{\omega},1\right)$, and then combining these variables through the following expression:
    \begin{equation}
        f\!\left(x\right) = \left[\omega\!\left(dx-\phi\right)\right]\,\mathrm{mod}\,\,\,1
    \end{equation}
    \item For the square wave dataset, each task's true function is produced by first sampling a wave frequency $\omega \sim \mathcal{U}\!\left(1,3\right)$ and a phase shift $\phi \sim \mathcal{U}\!\left(\frac{1}{\omega},1\right)$, and then combining these variables through the following expression:
    \begin{equation}
        f\!\left(x\right) =  \llbracket\left[\left\lfloor\omega x - \phi\right\rfloor\right]\,\mathrm{mod}\,\,\,2 = 0\rrbracket
    \end{equation}
    where $\left\lfloor\cdot\right\rfloor$ indicates the floor function, and $\llbracket P\rrbracket$ is the Iverson bracket of $P$, which takes on a value of 1 if the statement $P$ is true, and 0 otherwise.
    \item For the GP dataset, each task's true function is simply sampled from a Gaussian process with an Exponentiated Quadratic (EQ) kernel with lengthscale $\ell = 0.25$.
\end{itemize}

During training, the number of context points in each task is randomly selected to be any integer between 0 and 30. During testing, on the other hand, each of the three test meta-datasets utilised for benchmarking is composed of 310 tasks, evenly split across context sizes $\lvert\mathcal{D}_c\rvert$ comprised between 0 and 30 (i.e.~10 tasks for each). In all cases, the cardinality $\lvert\mathcal{D}_t\rvert$ of the original target set is fixed at 50. This setup allows for a highly granular analysis of DANP performance across varying context sizes (which, as evidenced by the results, proves crucial in drawing specific conclusions about the behaviour of different models). One should also note that, to ensure proper deployment, the AR ConvCNP model is trained on context sizes between 0 and 80 (where this upper limit arises from the sum of the maximum context set size of 30 and the target set size of 50).

\subsection{Model Architectures}

All models compared in this paper (including both the baselines and the DANPs)  utilise a U-Net architecture with 6 channels, each with size 64 and stride 2. Additionally, all models were trained for 500 epochs (each consisting of $2^{14}$ tasks), using a batch size of 16 and the Adam optimiser with a learning rate of 0.0003.

Lastly, three more hyperparameters must be specified for each of the three DANPs analysed above:

\begin{table}[h]
  \caption{DANP hyperparameters for each meta-dataset.}
  \label{tab:params}
  \centering
  \begin{tabular}{@{}P{3cm} P{3cm}P{3cm}P{3cm}@{}}
    \toprule
      & Base Model & Number of Noise Levels $F$ & Noise Variance $\beta$\\
    \midrule
    Sawtooth & ConvGNP & 3 & 0.08526\\
    \midrule
    Square Wave & ConvGNP & 3 & 0.153\\
    \midrule
    GP & ConvGNP & 2 & 0.2115\\
    \bottomrule
  \end{tabular}
\end{table}

The seemingly peculiar values of $\beta$ given above were selected in such a way as to yield, upon compounding of the noise through the layers, a pre-specified noise variance in the last fidelity level (0.02, 0.06 and 0.08 for sawtooth, square wave and GP respectively). More specifically, through simple algebra, one can show that the noising process described in the main text leads to the signal at the last fidelity $F$ taking on the following form:
\begin{equation}\label{eq:chap3_yF}
    y_F = \left(1-\beta\right)^{\frac{F}{2}}y_0 +\delta_F;\qquad \delta_F \sim \mathcal{N}\!\left(0, \beta - \beta\left(1-\beta\right)^F\right)
\end{equation}
which means that, given desired values for $F$ and for the noise variance $\sigma^2$ in fidelity level $F$, the appropriate value for $\beta$ can easily be obtained by numerically solving the following equation:
\begin{equation}
    \sigma^2 = \beta - \beta\left(1-\beta\right)^F
\end{equation}

\end{document}